\documentclass[sigconf]{acmart}

\usepackage{multirow}
\usepackage{algorithmic}
\usepackage{algorithm}
\AtBeginDocument{%
  \providecommand\BibTeX{{%
    \normalfont B\kern-0.5em{\scshape i\kern-0.25em b}\kern-0.8em\TeX}}}

\copyrightyear{2020} 
\acmYear{2020} 
\setcopyright{acmlicensed}\acmConference[KDD '20]{Proceedings of the 26th ACM SIGKDD Conference on Knowledge Discovery and Data Mining}{August 23--27, 2020}{Virtual Event, CA, USA}
\acmBooktitle{Proceedings of the 26th ACM SIGKDD Conference on Knowledge Discovery and Data Mining (KDD '20), August 23--27, 2020, Virtual Event, CA, USA}
\acmPrice{15.00}
\acmDOI{10.1145/3394486.3403088}
\acmISBN{978-1-4503-7998-4/20/08}

\usepackage{subcaption}


\begin{document}
\title{Policy-GNN: Aggregation Optimization \\ for Graph Neural Networks}
\author{Kwei-Herng Lai, Daochen Zha, Kaixiong Zhou, Xia Hu}
\email{{khlai037, daochen.zha, zkxiong, xiahu}@tamu.edu}
\affiliation{%
  \institution{Department of Computer Science and Engineering, Texas A\&M University}
  }

\begin{abstract}
Graph data are pervasive in many real-world applications. Recently, increasing attention has been paid on graph neural networks (GNNs), which aim to model the local graph structures and capture the hierarchical patterns by aggregating the information from neighbors with stackable network modules. Motivated by the observation that different nodes often require different iterations of aggregation to fully capture the structural information, in this paper, we propose to explicitly sample diverse iterations of aggregation for different nodes to boost the performance of GNNs. It is a challenging task to develop an effective aggregation strategy for each node, given complex graphs and sparse features. Moreover, it is not straightforward to derive an efficient algorithm since we need to feed the sampled nodes into different number of network layers. To address the above challenges, we propose \emph{Policy-GNN}, a meta-policy framework that models the sampling procedure and message passing of GNNs into a combined learning process. Specifically, Policy-GNN uses a meta-policy to adaptively determine the number of aggregations for each node. The meta-policy is trained with deep reinforcement learning~(RL) by exploiting the feedback from the model. We further introduce parameter sharing and a buffer mechanism to boost the training efficiency. Experimental results on three real-world benchmark datasets suggest that \emph{Policy-GNN} significantly outperforms the state-of-the-art alternatives, showing the promise in aggregation optimization for GNNs.
\end{abstract}

\maketitle

\section{Introduction}
Graph data are pervasive in many real-world applications, such as anomaly detection~\cite{LiHLDZ19}, molecular structure generation~\cite{NIPS2018_7877}, social network analysis~\cite{hu2013exploiting} and recommendation systems~\cite{he2017neural}. One of the key techniques behind these applications is graph representation learning~\cite{hamilton2017representation}, which aims at extracting information underlying the graph-structured data into low dimensional vector representations. Following the great success of convolution neural networks~\cite{lecun1998gradient}, a lot of momentum has been gained by graph convolution networks (GCNs)~\cite{KipfW17,DefferrardBV16} and graph neural networks (GNNs)~\cite{battaglia2018relational,zhou2018graph,you2018graphrnn,bresson2017residual,VelickovicCCRLB18} to model the local structures and hierarchical patterns of the graph.


Recently, many research efforts have been focused on pushing forward the performance boundary of GNNs. Previous studies of advancing GNNs mainly fall into two directions. First, sampling strategies such as batch sampling~\cite{HamiltonYL17} and importance sampling~\cite{ChenMX18} are proposed to improve learning efficiency. However, sampling strategy may cause information loss and thus delivers sub-optimal performance. Second, some novel message passing functions have been introduced to better capture the information underlying the graph structures, such as pooling layers~\cite{gao2019graph} and multi-head attention~\cite{VelickovicCCRLB18}. More recent work~\cite{oono2019asymptotic,gurel2019anatomy} has focused on deep architectures and studied the oversmoothing problem, which leads to gradient vanishing and makes features of vertices to have the same value~\cite{gurel2019anatomy}. One way to address the oversmoothing problem and make GNNs go deeper is to use skip-connection~\cite{abs-1904-03751,Huang0RH18,gao2019graph}. Although introducing skip-connection into GNNs achieves promising performance, this mechanism may be sub-optimal since it requires manually specifying the starting layer and the end layer for constructing the skip-connection. The predefined message passing architecture may not be suitable for all of the vertices in the graph.



\begin{figure}
    \centering
    \includegraphics[width=1.0\linewidth]{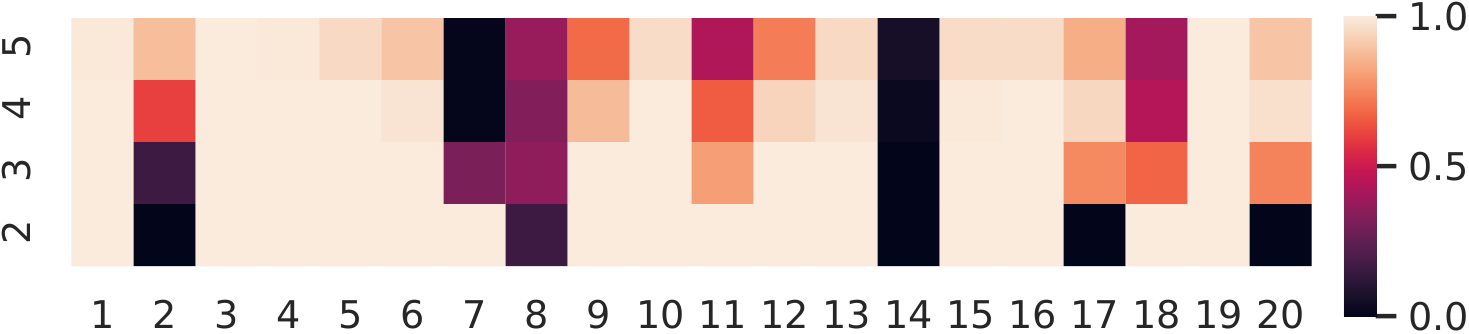}
    \caption{The effect of different iterations of aggregation in GCN on $20$ randomly sampled nodes of Cora dataset. The X-axis denotes the id of the sampled node, and Y-axis is the iteration number (layers) in GCN. The color from dark to light represents the ratio of being predicted correctly with 100 different runs.}
    \vspace{-10pt}
    \label{fig:prelim}
    \vspace{-10pt}
\end{figure}

In this work, we advocate explicitly sampling diverse iterations of aggregation for different nodes to achieve the best of both worlds. Specifically, we hypothesize that different nodes require different iterations of aggregation to fully capture the structural information. To understand our hypothesis, we apply standard GCN with different layers to conduct node classification on Cora dataset. Figure~\ref{fig:prelim} illustrates the ratio of being predicted correctly in $100$ runs of $20$ randomly sampled nodes. We observe that some nodes can achieve better classification performance with more GNN layers. While most of the nodes are classified well in 2 hop aggregation, some nodes such as node 17 and 20 only perform well when aggregating $4$ iterations, and some nodes such as node 2 requires more iterations of aggregation. The above observations motivate us to study how to adaptively aggregate different hops of neighbors for each node to boost the performance of GNNs.

However, this task is nontrivial due to the following challenges. First, real-world graphs are usually complex with multiple types of attributes: it is hard to determine the suitable iterations of aggregation for each node. For example, in the citation network such as Cora dataset, each node represents a piece of paper with 1,433 dimensions of bag-of-words features. It is difficult to design the aggregation strategy based on such sparse and large number of features. Second, even though we can define a proper aggregation strategy for each node, it remains challenging to train GNNs on these nodes since we need to feed these nodes into different number of network layers. Improper management of the sampled nodes will greatly affect the training efficiency of the algorithms.

To address the above challenges, in this paper, we propose Policy-GNN, a meta-policy framework to model the complex aggregation strategy. 
Motivated by the recent success of meta-policy learning~\cite{zha2019experience}, we formulate the graph representation learning as a Markov decision process (MDP), which optimizes a meta-policy by exploiting the feedback from the model. Specifically, the MDP iteratively samples the number of hops of the current nodes and the corresponding neighboring nodes with a meta-policy, and trains GNNs by aggregating the information of the nodes within the sampled hops. The proposed MDP successfully integrates the sampling procedure and message passing into a combined learning process. To solve this MDP, we employ deep reinforcement learning algorithms enhanced by a tailored reward function to train the meta-policy. In addition, we introduce parameter sharing and a buffer mechanism that enables batch training for boosting the training efficiency.

We showcase an instance of our framework by implementing it with deep Q-learning (DQN)~\cite{MnihKSRVBGRFOPB15} and GCN, which are the basic algorithms of reinforcement learning and GNNs, respectively. To validate the effectiveness of the proposed framework, we conduct extensive experiments on several real-world datasets against various state-of-the-art baselines, including network embedding methods, static GNN methods and graph neural architecture search methods. Experimental results suggest that the proposed framework achieves state-of-the-art performance on three benchmark datasets, i.e., Cora, CiteSeer and PubMed. The main contributions are summarized as follows:
\begin{itemize}
    \item We discover that introducing diverse aggregation to the nodes can significantly boost the performance of GNNs.
    \item We formulate the GNN training problem as a Markov decision process and propose a meta-policy framework that uses reinforcement learning to adaptively optimize the aggregation strategy.
    \item We develop a practical instance of our framework based on DQN and GCN with tailored reward function. The proposed algorithm significantly outperforms all the state-of-the-art alternatives that we are aware of, on the three real-world benchmark datasets. The code is made publicly available\footnote{\url{https://github.com/datamllab/Policy-GNN}}.
\end{itemize}



\begin{table}
\caption{Main notations in the paper. The top rows are for graph representation learning; the bottom rows cover deep reinforcement learning.}
\vspace{-5pt}
\label{tb:1}
\centering
\begin{tabular}{@{}ll@{}}
\toprule
\textit{Notation} & Definition        \\ \midrule
$V$             & The set of vertices in the graph.\\
$E$             & The set of edges in the graph.\\
$G$             & A graph G with node set $V$ and edge set $E$.\\
$\textbf{A}$    & The adjacency matrix of graph $G$.  \\
$\widetilde{\textbf{A}}$ & The normalized adjacency matrix of graph $G$ \\
$\textbf{X}$    & The attribute information matrix.  \\
$N_{k}(v)$      & The $k$-hop neighborhood of the node $v$. \\
\midrule
$\mathcal{S}$   & A finite set of state \\
$\mathcal{A}$   & A finite set of actions \\
$\mathcal{P}_{T}$ & State transition probability function. \\
$s_t$           & A state in the timestep $t$. \\
$a_t$           & An action in the timestep $t$. \\
$r_t$           & Reward in the timestep $t$. \\
$Q(s,a)$        & Value function to evaluate the state $s$ and action $a$. \\
$\widetilde{\pi}$ & Policy function to generate action $a$ for a given state $s$. \\
$b$ & The window size of the baseline in reward function. \\

\toprule
\end{tabular}
\vspace{-10pt}
\end{table}

\section{Preliminaries}
In this section, we first define the problem of aggregation optimization. Then we introduce the background of graph neural networks, Markov decision process, and deep reinforcement learning.

\subsection{Problem Definition}


Let $G=(V,E)$ denote a graph, where $V=\{v_1, v_2, v_3...v_n\}$ is the set of vertices, $E \subseteq V \times V$ is the set of edges and $n$ is the number of vertices in $G$. Each edge $e=(u,v,w) \in E$ consists of a starting node $u \in V$, an end node $v \in V$ and a edge weight $w \in \mathbb{R}^{+}$ that indicates the strength of the relation between $u$ and $v$. We use the convention of adjacency matrix $\textbf{A} \in \mathbb{R}^{n \times n}$ and attribute information matrix $\textbf{X} \in \mathbb{R}^{n \times m}$to represent $G$, where each entry $\textbf{A}_{ij} \in \mathbb{R}^{+}$ is the edge weight between $v_i$ and $v_j$, and each row $\textbf{X}_{i}$ is a $m$-dimensional attribute vector for node $v_i$. For each node $v$, we define a set of k-hop neighborhood as $N_k(v)$, where each $v' \in N_k(v)$ is the k-order proximity of the node $v$. Graph representation learning aims at embedding the graph into low-dimensional vector spaces. Formally, the objective can be expressed by a mapping function $\mathcal{F}: v \to \mathbb{R}^{d}$, where $d$ is the dimension of the vector. Following the message passing strategy in GNNs, our goal is to learn the node representation through aggregating the information from k-hop neighborhood $N_k(v)$. In this way, the proximity information is preserved in the low dimensional feature vectors and the learned representations can be used in down downstream tasks such as node classification. Table~\ref{tb:1} lists the main notations used in the paper.

Based on the notations defined above, we formulate the problem of aggregation optimization as follows. Given an arbitrary graph neural network algorithm, our goal is to jointly learn a meta-policy $\widetilde{\pi}$ with the graph neural network, where $\widetilde{\pi}$ maps each node $v \in V$ into the number of iterations of aggregation, such that the performance on downstream task is optimized.

\subsection{Learning Graph Representations with Deep Neural Networks}
\label{subsec:gnn}
Due to the great success of convolution neural networks on image data, graph neural networks~\cite{VelickovicCCRLB18,KipfW17,HamiltonYL17,DefferrardBV16} have been extensively studied for graph data. Inspired by Weisfeiler-Lehman (WL) graph isomorphism test~\cite{ShervashidzeSLMB11}, 
learning a graph neural network consists of 3 steps: (1) \textbf{initialization}: initialize the feature vectors of each node by the attribute of the vertices, (2) \textbf{neighborhood detection}: determiner the local neighborhood for each node to further gather the information and (3) \textbf{information aggregation}: update the node feature vectors by aggregating and compressing feature vectors of the detected neighbors.

To effectively gather the information from neighbors, several feature aggregation functions (graph convolution) have been proposed. One of the most representative method is \textbf{Graph Convolution Network (GCN)}~\cite{KipfW17}. Let $\widetilde{\textbf{A}}$ be the normalized adjacency matrix where $\widetilde{\textbf{A}}_{uv} = \frac{\textbf{A}_{uv}}{\sum_{v}\textbf{A}_{uv}}$, GCN performs weighted feature aggregation from the neighborhood: 
\begin{equation}
    \label{eq:gcn}
    \textbf{h}_{v}^{k} = \sigma(\sum\nolimits_{u \in \{v\} \cup N_{1}(v)} \widetilde{\textbf{A}}_{uv}\textbf{W}_{k}\textbf{h}_{u}^{k-1}).
\end{equation}
where $\textbf{W}_{k}$ is the trainable parameter in the k-th graph convolution layer, $N_1(v)$ is the one hop neighborhood of the node $v$, and $\textbf{h}_{u}$ and $\textbf{h}_{v}$ are $d$ dimensional feature vectors.


    
Recently, \textbf{Graph Attention Network (GAT)}~\cite{VelickovicCCRLB18} introduced the attention mechanism to indicate the importance of the neighbors of a given node on aggregation. GAT substitute $\widetilde{\textbf{A}}$ in equation~\ref{eq:gcn} with self-attention scores, which indicate the importance of aggregation from node $u$ to node $v$.

Another line of studies focus on improving the learning efficiency with sampling techniques. The original GCN performs aggregation for each node from all its directed neighbors at the same time, which becomes expensive when the size of graph gets larger. To alleviate the problem, there are mainly two kinds of sampling strategy. First, \textbf{Local Sampling} proposed by GraphSAGE~\cite{HamiltonYL17} performs aggregation for each node. Different from the original GCN, GraphSAGE only aggregates a fixed number of directed neighbors of each node. Second, \textbf{Global Sampling} is proposed by FastGCN~\cite{ChenMX18}. The idea of global sampling is introducing an importance sampling among all of the nodes, in order to build a new structure based on the topology of the original graph. Then the aggregation of the sampled neighbors will capture the global information of the graph.

In our work, inspired by the sampling approaches above, we formulate the local sampling into Markov decision process, and learn the global information through training a meta-policy to make decisions on k-hop aggregation. To make our contribution focused, we build our framework upon GCN, the basic form of GNNs. One can easily extend our framework to more advanced GNNs.


\begin{figure}
    \centering
    \includegraphics[width=0.95\linewidth]{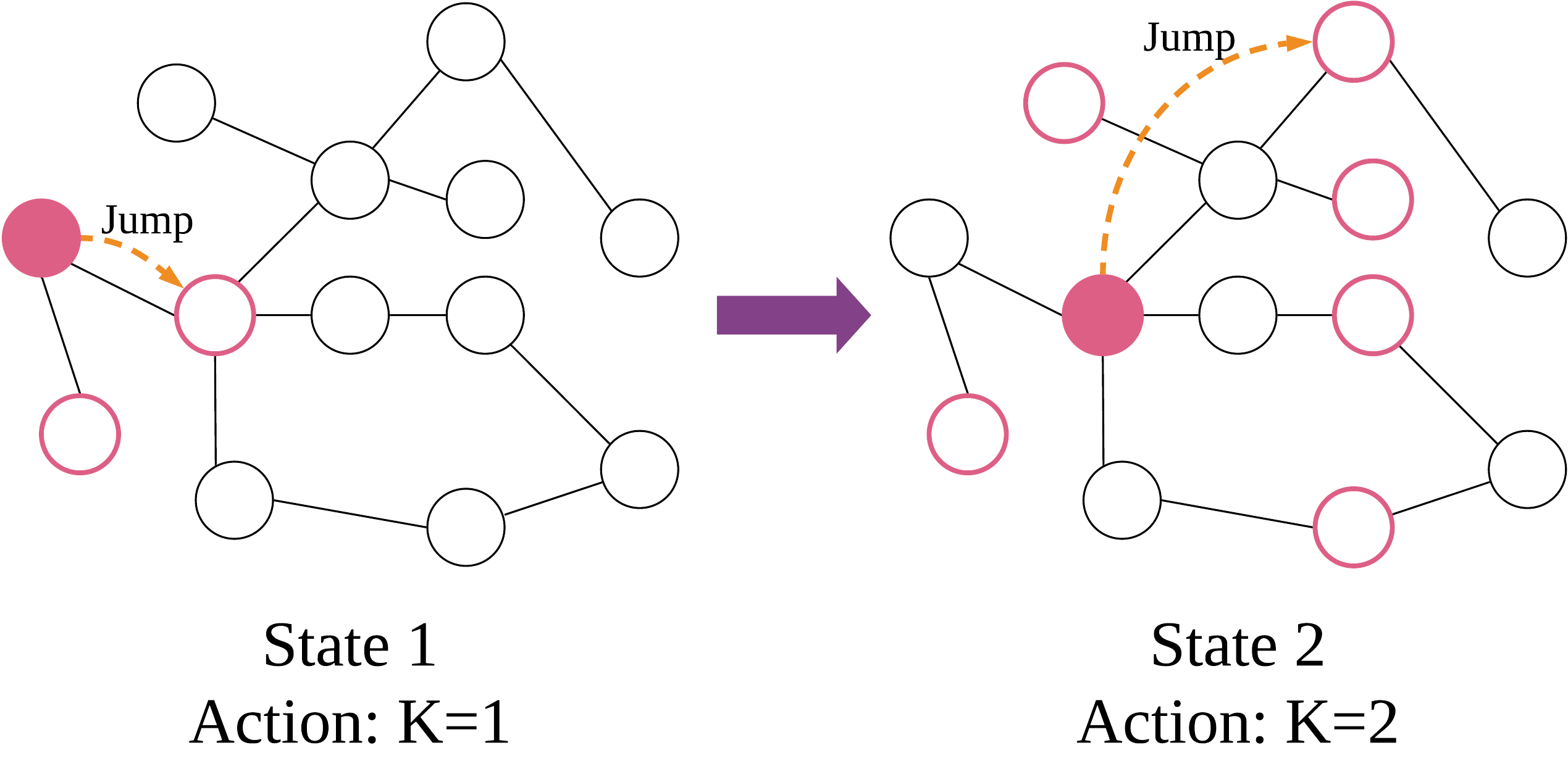}
    \caption{An illustration of node sampling as Markov decision process~(MDP). State 1 is the attribute of the solid-pink node in the left figure. We take action $K=1$ and randomly jump to one of $1$-hop neighboring nodes (pink-framed). As a result, state 1 transits to state 2, which is the attribute of the solid-pink node in the right figure. We then sample the next node from $2$-hop neighbors by taking action $K=2$. The above procedure can be naturally treated as an MDP.}
    \label{fig:MDP}
\end{figure}

\begin{figure*}
    \centering
    \includegraphics[width=0.9\linewidth]{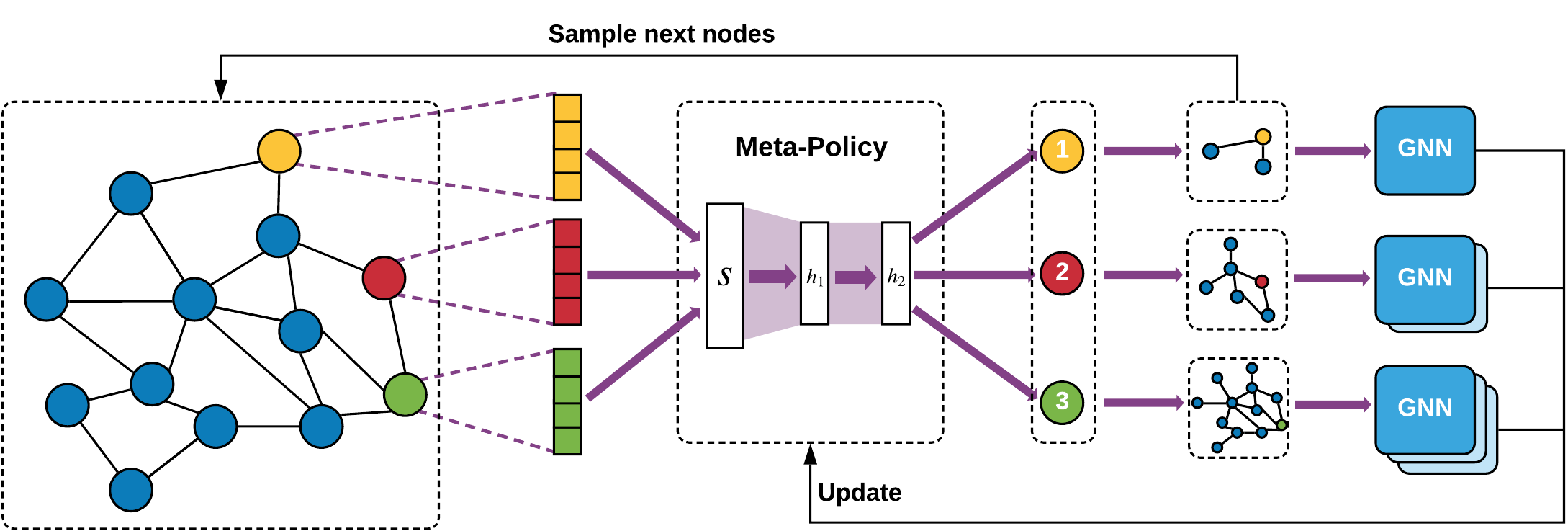}
    \caption{An illustration of Policy-GNN with meta-policy sampling $1$, $2$ and $3$ iterations of aggregation for different nodes. The learning procedure follows a Markov Decision Process. In each timestep, the mata-policy samples the number of layers (action) based on the attributes of the nodes (state). The next nodes (next state) are obtained by randomly sampling a node from the k-hop neighbors of the current nodes, where $k$ is the output of the mata-policy (action). The meta-policy is updated based on the feedback from the GNN.}
    \label{fig:overview}
\end{figure*}
\subsection{Sequential Decision Modeling with Markov Decision Process}
\label{subsec:markov}
Markov Decision Process~(MDP) is a mathematical framework to describe sequential decision making process. Specifically, let $\mathcal{M}$ be an MDP, represented by a quintuple $(\mathcal{S}, \mathcal{A}, \mathcal{P}_{T}, \mathcal{R}, \gamma)$, where $\mathcal{S}$ is a finite set of states, $\mathcal{A}$ is a finite set of actions, $\mathcal{P}_{T}: \mathcal{S} \times \mathcal{A} \times \mathcal{S} \to\mathbb{R}^+$ is the state transition probability function 
that maps the current state $s$, action $a$ and the next state $s'$ to a probability, $\mathcal{R}: \mathcal{S} \to \mathbb{R}$ is the immediate reward function, and $\gamma \in (0,1)$ is a discount factor. At each timestep $t$, the agent takes action $a_t \in \mathcal{A}$ based on the current state $s_t \in \mathcal{S}$, and observes the next state $s_{t+1}$ as well as a reward signal $r_t = \mathcal{R}(s_{t+1})$. We aim to search for the optimal decisions so as to maximize the expected discounted cumulative reward, i.e., we would like find a policy $\pi: \mathcal{S} \to \mathcal{A}$ to maximize $\mathbb{E}_\pi[\sum_{t=0}^{\infty}\gamma^t r_t]$. In the following sections, we discuss how graph representation learning can be formulate as a Markov decision process. Specifically, we describe the definitions of state, action and reward in the context of graph representation learning. 

\subsection{Solving Markov Decision Process with Deep Reinforcement Learning}
Deep reinforcement learning is a family of algorithms that solve the MDP with deep neural networks. In this work, we consider model-free deep reinforcement learning, which learns to take the optimal actions through exploration. The pioneering work Deep-Q Learning (DQN)~\cite{mnih2015human} uses deep neural networks to approximate state-action values Q(s, a) that satisfies
\begin{equation}
     \label{eq:qlearning}
      Q(s,a) = \mathbb{E}_{s'} [\mathcal{R}(s') + \gamma \max_{a'} (Q(s', a')],
\end{equation}
where $s'$ is the next state and $a'$ is the next action. DQN introduces two techniques to stabilize the training: (1) a replay buffer to reuse past experiences; (2) a separate target network that is periodically updated. In this work, we employ DQN, the most commonly used model-free algorithm, to solve The MDP; one could also employ more advanced algorithms in our framework.

\section{Methodology}
\label{sec:method}
Figure~\ref{fig:overview} illustrates an overview of our framework. There are two components in our framework, a \emph{Meta-Policy} module and a \emph{GNN} module. The \emph{Meta-Policy} aims at learning the relations between node attributes and the iterations of aggregation. \emph{GNN} module exploits the \emph{Meta-Policy} to learn the graph representations.

Through combining the two modules, we formulate the graph representation learning as a Markov decision process. Specifically, \emph{Meta-Policy} treats the node attributes as state (the red/yellow/green feature vectors), and maps the state into action (the number of hops shown in the red/yellow/green circles). Then, we sample the next state from the $k$-hop neighbors (the sub-graphs next to the action) of each node, where $k$ is the output of the \emph{Meta-Policy} (action). Thereafter, the \emph{GNN} module (at the right side of Figure~\ref{fig:overview}) selects a pre-built $k$-layer GNN architecture ($k$ is the output of the meta-policy) to learn the node representations and obtains a reward signal for updating the \emph{Meta-Policy}.

In what follows, we elaborate on the details of Policy-GNN. We first describe how we can train the meta-policy with deep reinforcement learning. Then, we show how we train GNNs with the meta-policy algorithm. Last, we introduce the buffer mechanism and parameter sharing strategy, which boost the training efficiency in real-world application scenarios.

\subsection{Aggregation Optimization with Deep Reinforcement Learning}
\label{subsec:mdp}
We discuss how the process of learning an optimal aggregation strategy can be naturally formulated as a Markov decision process (MDP). As discussed in Section~\ref{subsec:markov}, the key components of an MDP include states, actions and rewards, as well as a transition probability that maps the current state and action into the next state. With the above in mind, we now discuss how we define these components in the context of graph representation learning:



\begin{itemize}
    \item \textbf{State ($\mathcal{S}$)}: The state $s_t \in \mathcal{S}$ in timestep $t$ is defined as the attribute of the current node.
    
    \item \textbf{Action ($\mathcal{A}$)}: The action $a_t \in \mathcal{A}$ in timestep $t$ specifies the number of hops of the current node.
    
    \item \textbf{Reward Function($\mathcal{R}$)}: We define the reward $r_t$ in timestep $t$ as the performance improvement on the specific task comparing with the last state.
\end{itemize}

Based on the definitions above, the proposed aggregation process consists of three phases: 1) selecting a start node and obtaining its attributes as the current state $s_t$, 2) generating an action $a_t$ from $\pi(s_t)$ to specify the number of hops for the current node, and 3) sampling the next node from $a_t$-hop neighborhood and obtaining its attributes as the next state $s_{t+1}$. Figure~\ref{fig:MDP} gives an simple example of how does MDP work for graph representation learning. Specifically, we formulate the sampling process of GNN into a MDP. With the specified action $k$, we sampled the next state from the $k$-hop neighborhood.


We propose to employ deep reinforcement learning algorithms to optimize the above MDP. Since the action space of the aggregation process is a discrete space, i.e., $\mathcal{A} \in \mathbb{N}$ and any arbitrary action $a \in \mathcal{A}$ is always a finite positive integer, we introduce the deep Q-learning~\cite{MnihKSGAWR13,MnihKSRVBGRFOPB15} to address the problem.

The key factor in guiding the deep Q-learning is the reward signal. We employ a baseline in the reward function, defined as
\begin{equation}
\label{eq:rfunc}
\mathcal{R}(s_{t}, a_{t}) =\lambda (\mathcal{M}(s_{t}, a_{t}) - \frac{\sum_{i=t-b}^{t-1}\mathcal{M}(s_{i}, a_{i})}{b-1}),
\end{equation}
where $\frac{\sum_{i=t-b}^{t-1}\mathcal{M}(s_{i}, a_{i})}{b-1}$ is the baseline for each timestep $t$, $\mathcal{M}$ is the evaluation metric of a specific task, $b$ is a hyperparameter to define the window size for the historical performance to be referred for the baseline, $\lambda$ is a hyperparameter to decide the strength of reward signal and $\mathcal{M}(s_{t}, a_{t})$ is the performance of the task on the validation set at timestep $t$. The idea of employing a baseline function is to encourage the learning agent to achieve better performance compared with the most recent $b$ timesteps. In this work, we showcase the learning framework by using the node classifications accuracy on the validation set as the evaluation metric.

Based on the above reward function, we train the $Q$ function by optimizing Equation~(\ref{eq:qlearning}). In this way, we can apply epsilon-greedy policy~\cite{Watkins:1989} to obtain our policy function $\widetilde{\pi}$.
\begin{algorithm}[t]
\caption{Policy-GNN}
\label{alg:overview}
\setlength{\intextsep}{0pt} 
\begin{algorithmic}[1]
\STATE \textbf{Input:} Maximum layer number $K$, DQN training step $S$, total training epoch $T$, epsilon probability $\epsilon$, window size of reward function $b$.
\STATE Initialize $K$ GNN layers, $Q$ function, memory buffer $\mathcal{D}$, GNN buffer $\mathcal{B}$
\STATE Randomly sample a node, generate a state $s_{0}$ by attribute of the nodes.
\FOR{$t$ = $0, 1, 2 ..., T$}
    \STATE With probability $\epsilon$ randomly choose an action $a_t$, \\               otherwise obtain $a_t=\mathrm{argmax}_{a}Q(s_t, a)$.
    \STATE Store $s_t$ and $a_t$ into GNN buffer $\mathcal{B}_{a_t}$
    \STATE Apply Algorithm~\ref{alg:buffer} with input $a$ and $\mathcal{B}$ to train the GNNs.
    \STATE Obtain $r_t$ on validation dataset via equation~\ref{eq:rfunc}
    \STATE Sample the next state $s_{t+1}$ from ${a_t}$-hop neighborhood.
    \STATE Store the triplet $T_{t} = (s_{t}, a_{t}, s_{t+1}, r_{t})$ into $\mathcal{D}$
    \FOR{step = $1$, $2$, .., $S$}
        \STATE Optimize $Q$ function using the data in $\mathcal{D}$ via equation~\ref{eq:qlearning}.
    \ENDFOR
\ENDFOR
\end{algorithmic}
\end{algorithm}

\subsection{Graph Representation Learning with Meta-Policy}
\label{subsec:grl}
We discuss the role that meta-policy plays in graph representation learning and how to learn the node embeddings with \emph{Meta-Policy}. In the proposed framework, we explicitly learn the aggregation strategy through the aforementioned meta-policy learning. As Figure~\ref{fig:overview} shows, the \emph{Meta-Policy} maps the node attributes to the number of hops, and uses the number of hops to construct a specific GNN architecture for each node. Compared with the recent efforts~\cite{abs-1904-03751,Huang0RH18} on skip-connection, which implicitly learns an aggregation strategy for the whole graph, the proposed framework explicitly aggregates the information.
Furthermore, the Policy-GNN provides the flexibility to include all of the previous research fruits, including skip-connection, to facilitate graph representation learning.

To showcase the effectiveness of the framework, we apply the meta-policy on basic GCN~\cite{KipfW17}, and learn the node representations by equation~\ref{eq:gcn}. Instead of aggregating a fixed number of layers for every nodes, our framework aggregates $a_t$ layers for each node attribute $s_t$ at timestep $t$. The GNN architecture construction can be represented as the transition equations as follows:
\begin{align*}
    &\textbf{h}_{v}^{1} = \sigma(\sum\nolimits_{u_1 \in \{v\} \cup N_{1}(v)}\widetilde{a}_{u_{1}v}\textbf{W}_{1}\textbf{X}_{u}), \\
    &\vdots \\ 
    &\textbf{h}_{v}^{k=a_t} = \sigma(\sum\nolimits_{u_{k} \in \{u_{k-1}\} \cup N_{k}(v)}\widetilde{a}_{u_{k}u_{k-1}}\textbf{W}_{k}\textbf{h}_{v}^{k-1}), \\
    & output =  \mathrm{softmax}(\textbf{h}_{v}^{a_t}),
\end{align*}
where $\textbf{h}_{v}^{k}$ is the $d$ dimensional feature vector of node $v$ of the $k$ layer, $\textbf{X}_{u}$ is the input attribute vector of node $u$, $\textbf{W}_{k}$ is the trainable parameters of the $k$ layer, $\sigma$ is the activation function of each layer, and $k=a_t$ is the number of aggregation that decided by $\widetilde{\pi}$ function in timestep $t$. Note that, we can always replace the aggregation function in each layer as well as the final output layer for different tasks in our framework.

\begin{algorithm}[t]
\caption{Training GNN with Buffer Mechanism}
\label{alg:buffer}
\setlength{\intextsep}{0pt} 
\begin{algorithmic}[1]
\STATE \textbf{Input:} Action $a$, GNN buffer $\mathcal{B}$

\IF{$\mathcal{B}_a$ is full}
    \FOR{layers = $1$, $2$, ... $a$}
        \STATE Stack GNN layers as subsection~\ref{subsec:grl} mentioned.
    \ENDFOR
    \STATE Train the stacked GNNs on the buffer of action $\mathcal{B}_a$
    \STATE Clear the buffer $\mathcal{B}_a$
\ENDIF
\end{algorithmic}
\end{algorithm}

\subsection{Accelerating Policy-GNN with Buffer Mechanism and Parameter Sharing}
One of the challenging problems to practically develop the proposed framework is training efficiency. Since re-constructing GNNs at every timestep is time-consuming, and the number of parameters may be large if we train multiple GNNs for different actions. To address the problems, we introduce the following techniques to improve training efficiency.

\textbf{Parameter Sharing.} We reduce the number of training parameters via parameter sharing mechanism. Specifically, we first initialize the maximum layer number ($k$) of GNN layers in the beginning. Then, in each timestep $t$, we repeatedly stack the layers by the initialization order for the action $a_t$ to perform GNN training. In this way, if the number of hidden units for each layer is $n$, we only need to train $k \times n$ parameters, rather than $\frac{nk(k+1)}{2}$ parameters.


\textbf{Buffer mechanism for graph representation learning.} We construct action buffers for each possible action. In each timestep, after we obtain a batch of nodes' attributes and a batch of actions, we store them into the action buffer and check if the buffer has reached the batch size. If the buffer of a specific action is full, then we construct the GNN with selected layers, and train the GNN with the data in the buffer of the action. After the data has been trained on the GNN, we clear the data in the buffer. Practically, if the batch size is too large, the mechanism will still take much time. However, compared with the costly GNN construction in every timestep for training only one instance, the mechanism significantly improves the training efficiency~\footnote{The preliminary result on comparing the training time with/without the mechanisms on Cora dataset shows that the training efficiency is significantly improved by 96 times.}. 

The detail of the algorithm is provided in algorithm~\ref{alg:overview} and~\ref{alg:buffer}. We first initialize $k$ GNN layers, the $Q$ function, the GNN buffer $\mathcal{B}$ and the memory buffer $\mathcal{D}$ to store the experiences. Then, we randomly sample a batch of nodes and generate the first state with the node attributes. We feed the state to $Q$ function and obtain the actions with $\epsilon$ probability to randomly choose the action. Based on the chosen action, we stack the layers from the initialized GNN layers in order, and train the selected layers to get the feedback on the validation set. With the feedback on the validation set, we further obtain the reward signal via equation~\ref{eq:rfunc} and store the transitions into a memory buffer. In this way, with past experiences in the memory buffer, we randomly sample batches of data from the memory buffer and optimize the $Q$ function based on equation~\ref{eq:qlearning} for the next iteration.

\section{Experiment}
In this section, we empirically evaluate the proposed \emph{Policy-GNN} framework. We mainly focus on the following research questions:
\begin{itemize}
    \item How does \emph{Policy-GNN} compare against the state-of-the-art graph representation learning algorithms~(Section~\ref{sec:performance})?
 
    \item What does the distribution of the number of layers look like for a trained meta-policy~(Section~\ref{sec:layers})?
   \item Is the search process of \emph{Policy-GNN} efficient in terms of the number of iterations it needs to achieve good performance~(Section~\ref{sec:search})?
\end{itemize}

\subsection{Datasets}
\label{subsec:data}
We consider the benchmark datasets~\cite{sen2008collective} that are commonly used for studying node classification, including the citation graph-structured data of Cora, Citeseer, and Pubmed. The statistics of these datasets are summarized in Table~\ref{tab:dataset}. In these citation graphs, the nodes and the edges correspond to the published documents and their citation relations, respectively. The node attribute information is given by bag-of-words representation of a document, and each node is associated with a class label. We follow the standard training/validation/test splits as in previous studies~\cite{VelickovicCCRLB18,gao2018large, zhou2019auto,gao2019graphnas} 
Specifically, we use $20$ nodes per class for training, $500$ nodes for validation, and $1000$ nodes for testing. The proposed model and baselines are all trained and evaluated with the complete graph structure and node features in the training dataset, without using the node labels in the held-out validation and testing sets. The model hyperparameters are selected based on the performance on the validation set. The final classification accuracy is reported on the test set. 

\vspace{-5pt}
\subsection{Baselines}
We compare \emph{Policy-GNN} with the state-of-the-art graph representation learning algorithms. Specifically, we consider the baselines in the following three categories: network embedding methods, traditional graph neural networks (Static-GNNs), and the recently proposed neural architecture search based GNNs (NAS-GNNs). Note that NAS-GNNs also include a meta-policy that operates upon the GNNs. The search space of NAS-GGNs focuses on the message passing functions, whereas our meta-policy searches for the optimal iterations of aggregation. We are interested in studying whether our search strategy is more effective than NAS.

 \textbf{Network Embedding.} Network embedding learns the node representations in an unsupervised fashion. To capture the proximity information, network embedding utilizes the learning objective of the skip-gram method~\cite{mikolov2013distributed} and introduces various sampling techniques such as random walk. We consider \emph{DeepWalk}~\cite{PerozziAS14} and \emph{Node2vec}~\cite{grover2016node2vec}, which are the most commonly used baselines in network embedding. Both methods learn node feature vectors via random walk sampling. Compared with \emph{DeepWalk}, \emph{Node2vec} introduces a biased random walking strategy to sample the neighborhood for each node in BFS and DFS fashion.

\begin{table}[]
\setlength{\tabcolsep}{14pt}
  \caption{Summary of data statistics used in our experiment, including Cora, Citeseer, and Pubmed \cite{sen2008collective}.}
  \small
  \label{tab:dataset}
  \begin{tabular}{c|ccc}
    \toprule
     Dataset & Cora & Citeseer & Pubmed\\
    \midrule
    \midrule
    \#Nodes & $2708$ & $3327$ & $19717$ \\
    \#Features & $1433$ & $3703$ & $500$ \\
    \#Classes& $7$ & $6$ & $3$  \\
    \#Training Nodes & $140$ & $120$ & $60$ \\
    \#Validation Nodes & $500$ & $500$ & $500$\\
    \#Testing Nodes & $1000$ & $1000$ & $1000$ \\
  \bottomrule
\end{tabular}
\vspace{-15pt}
\end{table}

\begin{table*}[]
\caption{Classification accuracy on benchmark datasets Cora, Citeseer adn Pubmed. Symbol $\uparrow$ denotes the performance improvement achieved by \emph{Policy-GNN}, compared with network embedding, Static-GNNs, NAS-GNNs and random policy.}
\label{tbl:3}
\setlength{\tabcolsep}{7pt}
\begin{tabular}{l|c|c|cc|cc|cc}
\toprule
\multirow{2}{*}{\textbf{Baseline Class}} & \multirow{2}{*}{\textbf{Model}} & \multirow{2}{*}{\textbf{\#Layers}} & \multicolumn{2}{c|}{\textbf{Cora}} & \multicolumn{2}{c|}{\textbf{Citeseer}} & \multicolumn{2}{c}{\textbf{Pubmed}}  \\
\cline{4-9}
&        &         &     Accuracy   &  $\uparrow$    &  Accuracy &  $\uparrow$  &     Accuracy &  $\uparrow$    \\
\midrule
\midrule
\textbf{Network}
 &   DeepWalk~\cite{PerozziAS14}      &  -        &  0.672    &   +36.8\%      &    0.432    &   +107.6\%   & 0.653 & +41.0\%  \\
\textbf{Embedding} &  Node2vec~\cite{grover2016node2vec}  &  -        &  0.749   &   +22.7\%         &    0.547   &   +64.0\%    & 0.753 & +22.3\%   \\
\midrule
\multirow{7}{*}{\textbf{Static-GNNs}} 
 &   Chebyshev~\cite{DefferrardBV16}      &  2        &  0.812 &   +13.2\%           &  0.698 &  +28.5\%    & 0.744 &  +23.8\%    \\
 &       GCN~\cite{KipfW17}        &  2        &  0.815    &  +12.8\%         &  0.703  &  +27.6\%   & 0.790 &  +16.6\%    \\
 &     GraphSAGE~\cite{HamiltonYL17}    &  2        &  0.822      &  +11.8\%       &  0.714  & +25.6\%    & 0.871  & +5.7\%  \\
 &     FastGCN~\cite{chen2018fastgcn}      &   2       &  0.850     &   +8.1\%       &  0.776  & +15.6\%    & 0.880 &   +4.7\%   \\
 &      GAT~\cite{VelickovicCCRLB18}         &  2        &  0.830$\pm$ 0.006 & +10.7\%  &  0.725$\pm$ 0.005  &  23.7\%  & 0.790$\pm$ 0.002  &  +16.6\% \\
 &     LGCN~\cite{gao2018large}         &  2     &  0.833$\pm$ 0.004  & +10.3\%  &  0.730$\pm$ 0.004  &  +22.9\%  & 0.795$\pm$ 0.002  & +15.8\%  \\
 &     g-U-Nets~\cite{gao2019graph}     &  4        &  0.844$\pm$ 0.005 &  +8.9\%  & 0.732$\pm$0.003 &  +22.5\%  & 0.796$\pm$ 0.002  &  +15.7\%  \\
  &     Adapt~\cite{Huang0RH18}        &   2       &  0.874$\pm$ 0.003 &  +5.1\%  &  0.796$\pm$ 0.002   &  +12.7\%  & 0.906$\pm$ 0.002  & +1.7\%  \\
\midrule
\multirow{2}{*}{\textbf{NAS-GNNs}} 
 &    GraphNAS~\cite{gao2019graphnas} &  2 &  0.833 $\pm$ 0.002 &  +10.3\%  &  0.735$\pm$ 0.007 & +22.0\%  & 0.781$\pm$ 0.003 & +17.9\%  \\
 &    AGNN~\cite{zhou2019auto}     &  2 &  0.836 $\pm$ 0.003 & +9.9\%   &  0.738$\pm$ 0.005 & +21.5\%   & 0.797$\pm$ 0.003 &  +15.6\%  \\
\midrule
\multirow{2}{*}{\textbf{Policy-GNN}} 
& Random Policy            & 2 $\sim$ 5 & 0.770$\pm$ 0.021 & +19.4\%  & 0.656$\pm$ 0.024 & +36.7\%  & 0.788 $\pm$ 0.011  &  +16.9\% \\
& Policy-GNN     & 2 $\sim$ 5 & \textbf{0.919$\pm$ 0.014} & - & \textbf{0.897$\pm$ 0.021} & -   &  \textbf{0.921$\pm$ 0.022} & -  \\
\bottomrule
\end{tabular}
\end{table*}
    
\textbf{Static-GNNs.} GNNs learn the node representations in a supervised fashion, where neural networks are trained with both feature vectors and the labels in the training set. There are a huge number of algorithms in the literature. To better understand the performance of our \emph{Policy-GNN}, we try to include all the state-of-the-art methods that we are aware of for comparison. Specifically, we consider the following baselines which use different layer structure and batch sampling techniques. \emph{Chebyshev}~\cite{DefferrardBV16} and \emph{GCN}~\cite{KipfW17} perform information aggregation based on the Laplacian or adjacent matrix of the complete graph. \emph{GraphSAGE}~\cite{hamilton2017inductive} proposes neighborhood batch sampling to enable scalable training with max, min and LSTM aggregation functions. \emph{FastGCN}~\cite{ChenMX18} utilizes importance sampling in each aggregation layer to learn the feature vectors efficiently. \emph{GAT}~\cite{VelickovicCCRLB18} introduces multi-head attention mechanism into aggregation function, which learns the importance of the neighborhood of each node for information aggregation. \emph{LGCN}~\cite{gao2018large} automatically selects a fixed number of neighbors for each node based on value ranking, and transforms graph data into grid-like structures in 1-D format to extract the proximity information. \emph{Adapt}~\cite{Huang0RH18} builds up the network layer by layer in a top-down manner, where the nodes in the lower layer are sampled conditionally based on the upper layer. 
\emph{g-U-Nets}~\cite{gao2019graph} proposes an encoder-decoder structure with gPool layer to adaptively sample nodes to form a smaller graph, and a gUnpool to restore the original graph from the smaller graph and use skip-connection for deeper architectures.

\textbf{NAS-GNN.} Recently, neural architecture search has been extensively introduced to search the optimal GNN architectures within fix iterations of aggregation.
\emph{GraphNAS}~\cite{gao2019graphnas} uses a recurrent neural network to sample the variable-length architecture strings for GNNs, and applies reinforcement rule to update the controller. \emph{AGNN}~\cite{zhou2019auto} follows the setting in GraphNAS, and further introduces the constrained parameter sharing and reinforced conservative controller to explore well-performing GNNs efficiently. 
The search space of both baselines covers the aggregation functions, activation functions, attention functions, and the number of heads for multi-head attention mechanism. 

In addition to the above three categories, we also include a variant of our \emph{Policy-GNN}, named \emph{Random Policy}. \emph{Random Policy} will randomly make the decisions for each node without the reinforcement learning update. This baseline can be regarded as a special case of our framework epsilon probability $1.0$.

\vspace{-5pt}
\subsection{Implementation Details}
In the experiments, we implement our framework using \emph{GCN}~\cite{KipfW17} and \emph{DQN}~\cite{MnihKSGAWR13,MnihKSRVBGRFOPB15}. For \emph{GCN} layers, we apply \emph{relu} as the activation function for each layer and include the dropout mechanism between the layers with $0.5$ dropout rate. To train the GCN, we utilize \emph{Adam} optimizer with a learning rate $0.01$ and a weight decay rate $0.0005$, and set the batch size as $128$. We use the \emph{DQN} implementation in~\cite{zha2019rlcard} and construct 5-layers of \emph{MLP} with $(32,64,128,64,32)$ hidden units for $Q$ function. We set the memory buffer size to be $10000$ and the training batch size to be $256$. For the epsilon probability of policy function, we set up a linear scheduler with starting probability $1.0$ to the end probability $0.1$, where the probability linearly decades every $10$ training steps. We follow the setting of GraphNAS and AGNN by training the meta-policy for 1000 episodes. We save the model that has the best accuracy on the validation set and report the performance of the model by applying it on the test set. More details are available in Appendix~\ref{appendix}.

\vspace{-5pt}
\subsection{Performance Comparison on Benchmark Datasets}
\label{sec:performance}
Table~\ref{tbl:3} summarizes the performance of \emph{Policy-GNN} and the baselines on the three benchmark datasets. For our \emph{Policy-GNN}, the number of layers for each node is sampled from $2$ to $5$. The performance on the three datasets is measured by classification accuracy on the test data. We observe that the proposed \emph{Policy-GNN} significantly and consistently outperforms all of the baselines that we considered across all the three datasets. Compared with the second-best algorithm, \emph{Policy-GNN} improves the accuracy by $5.1\%$, $12.7\%$ and $1.7\%$ on Cora, Citeseer and Pubmed, respectively. We also make the following observations from the experimental results. 

First, the end-to-end learning procedure is superior to a separate pipeline. Specifically, compared with the network embedding approaches, all of the GNN approaches, including our framework, achieve much better performance. It is reasonable since network embedding approaches are unsupervised with two separate components of random walk sampling and skip-gram objective, which may be sub-optimal in node classification tasks.

\begin{figure}
    \centering
    \includegraphics[width=0.85\linewidth]{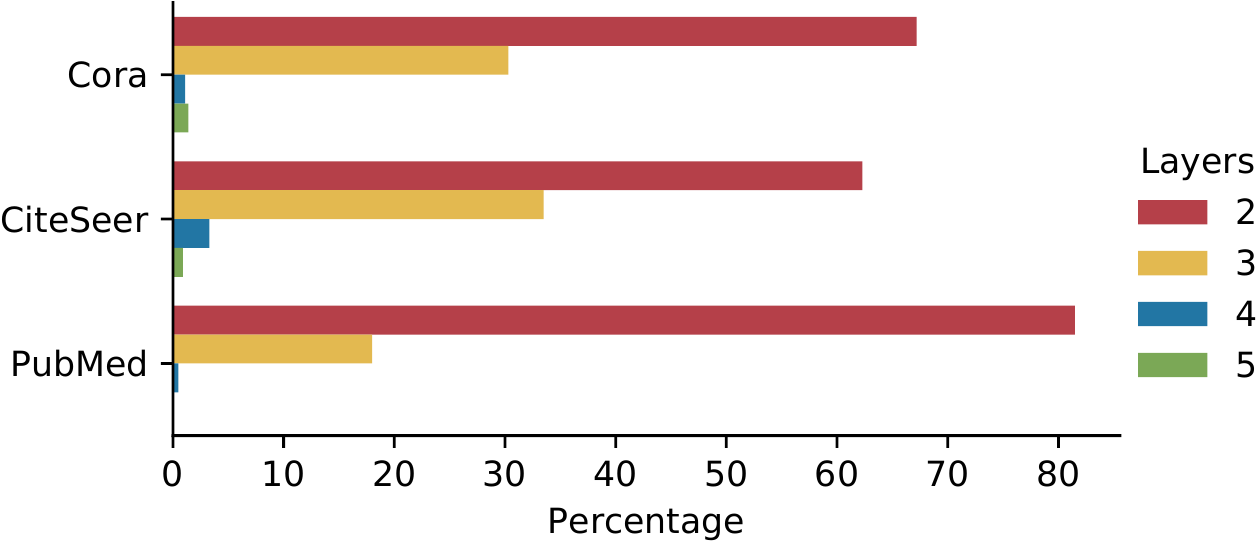}
    \vspace{-9pt}
    \caption{The percentage of the nodes that are assigned to different layers of GCN by \emph{Policy-GNN}. }
    \vspace{-8pt}
    \label{fig:dist}
\end{figure}

Second, using different iterations of aggregation for different nodes is critical for boosting the performance of GNNs. Specifically, built on \emph{GCN}, \emph{Policy-GNN} is able to significantly improve the vanilla \emph{GCN} by $12.8\%$, $27.6\%$ and $16.6\%$ on the three datasets, respectively. \emph{Policy-GNN} also outperforms the other more advanced Static-GNNs approaches. For example, our \emph{Policy-GNN} consistently outperforms \emph{Adapt}, which uses advanced important sampling techniques, in a large margin. Compared with \emph{g-U-Nets}, which applies skip-connection to deepen the architecture for the graph, our \emph{Policy-GNN} also achieves much better performance. The result suggests that the meta-policy successfully models the sampling procedure and message passing under a joint learning framework.


\begin{figure*}[t]
  \centering
    \begin{subfigure}[b]{0.33\textwidth}
    \centering
    \includegraphics[width=0.8\textwidth]{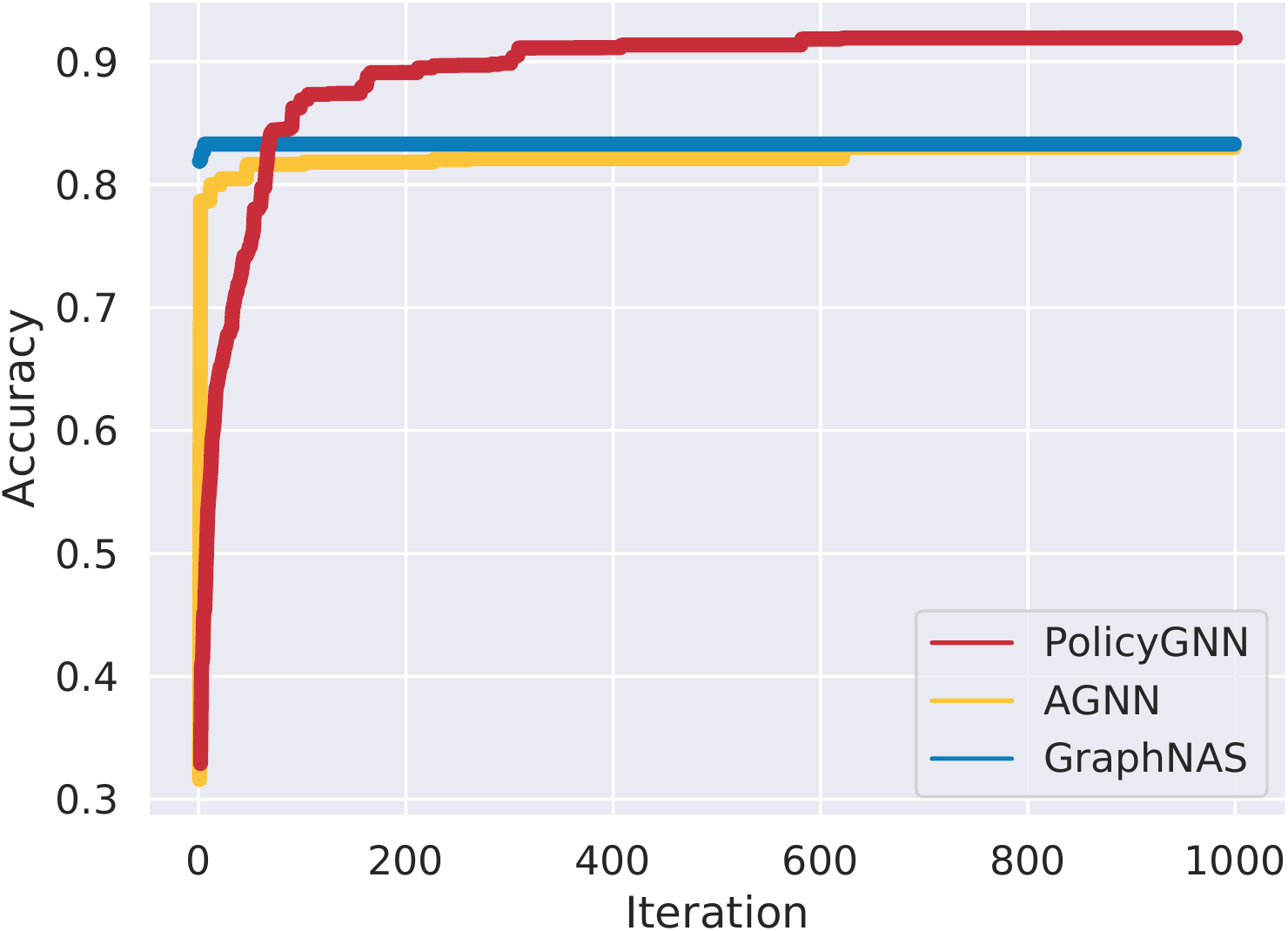}
    \caption{Cora}
    \label{fig:4-a}
  \end{subfigure}%
  \begin{subfigure}[b]{0.33\textwidth}
    \centering
    \includegraphics[width=0.8\textwidth]{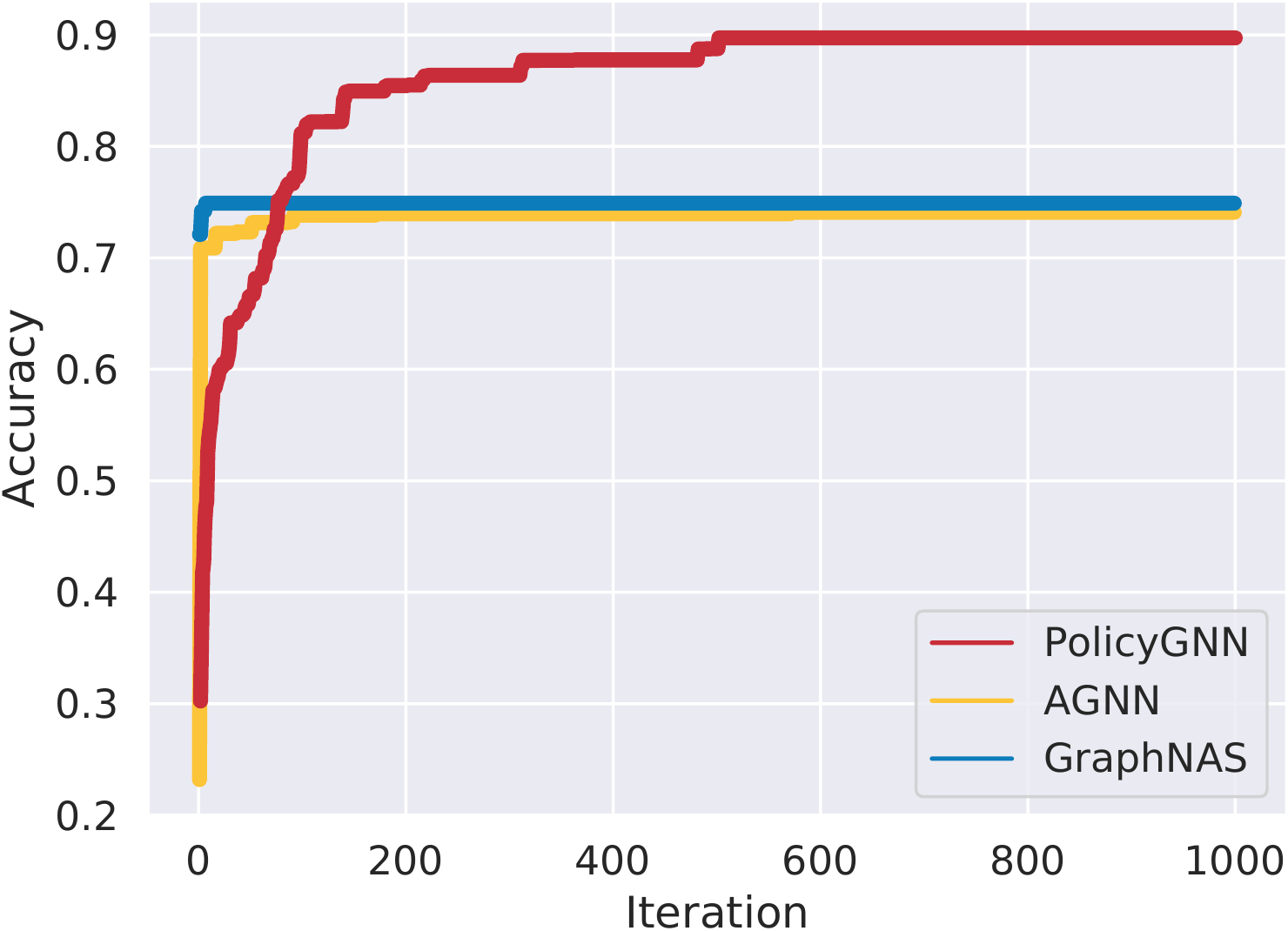}
    \caption{CiteSeer}
    \label{fig:4-b}
  \end{subfigure}%
  \begin{subfigure}[b]{0.33\textwidth}
    \centering
    \includegraphics[width=0.8\textwidth]{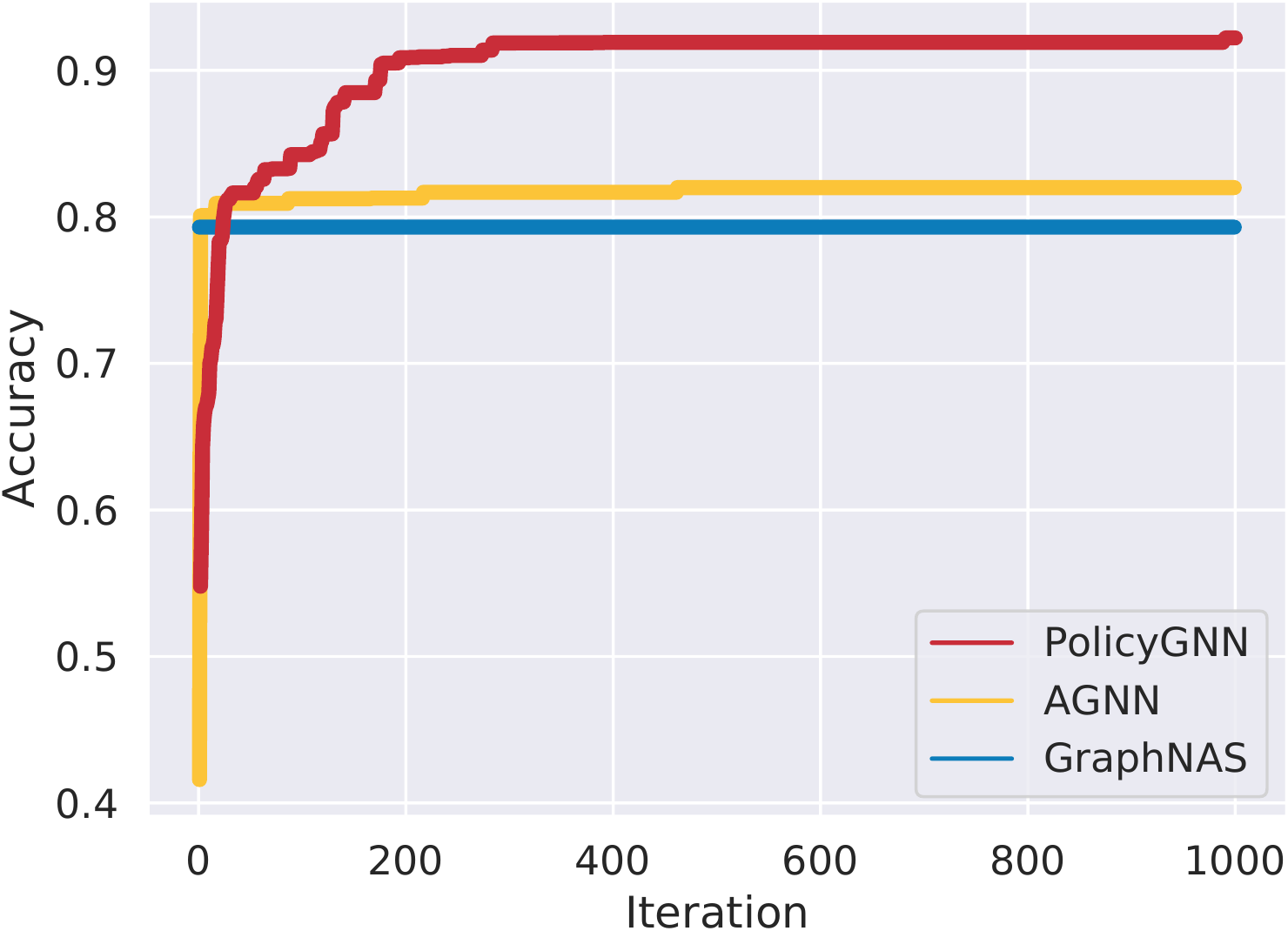}
    \caption{PubMed}
    \label{fig:4-c}
  \end{subfigure}%
  \caption{Learning curves of $1000$ iterations. We plot the accuracy with respect to the number of iterations on the three datasets. Although \emph{Policy-GNN} converges slightly slower than NAS methods, it achieves much higher accuracy later.}
  \label{fig:4}
\end{figure*}

Third, learning to sample different iterations of aggregation for different nodes is more effective than searching the neural architectures. Specifically, with only the basic \emph{GCN} layers, our \emph{Policy-GNN} is able to achieve state-of-the-art results by learning a proper aggregation strategy. Whereas, \emph{GraphNAS} and \emph{AGNN} only achieve marginal improvement over \emph{GCN} by searching the neural architectures. The possible reason is that, NAS methods require manually defined search space with the same number of layers setting for all nodes. This makes the discovered architecture limited by the search space, and therefore perform similar to static-GNNs on all three datasets. Thus, we argue that more research efforts should be made on aggregation strategy since proper iterations of aggregation will significantly boost the performance.


Last but not least, the result of \emph{Random Policy} demonstrates that reinforcement learning is a promising way to train the meta-policy in \emph{Policy-GNN}. Specifically, we observe that the performance of \emph{Random Policy} is significantly lower than \emph{Policy-GNN} and most of the baselines. This suggests that the proposed \emph{Policy-GNN} indeed learns an effective meta-policy for aggregation.

\subsection{Analysis of the Sampled Layers}
\label{sec:layers}

To better understand the learned meta-policy, in Figure~\ref{fig:dist}, we visualize what decisions have been made by the meta-policy. Specifically, we show the percentage of the nodes that are assigned to different layers of \emph{GCN}. As expected, most of the nodes only need two layers of aggregation. However, there are still some nodes that require three or more layers of aggregation. For example, in CiteSeer, more than $30\%$ of nodes requires $3$ iterations of aggregation and $5\%$ of nodes even requires $4$ iterations of aggregation. We also observe that the distributions of the iteration numbers are diverse in different datasets. This again verifies the necessity of using a learning-based meta-policy to model the diverse data characteristics.

\vspace{4pt}
\subsection{Analysis of the Learning Process}
\label{sec:search}
Figure~\ref{fig:4} plots the classification accuracy with respect to the number of iterations of our \emph{Policy-GNN} and the two neural architecture search baselines. We observe that \emph{Policy-GNN} learns slightly slower than \emph{AGNN} and \emph{GraphNAS} at the beginning. A possible explanation is that searching the optimal iterations for the nodes is a more challenging task than searching the neural architectures. We also observe that \emph{Policy-GNN} converges to much better accuracy than the NAS baselines in the later stages. The results suggest that searching for the aggregation strategy is indeed beneficial to GNNs.


\section{Related Work}
This section introduces previous studies on graph neural networks, graph neural architecture search and meta-policy learning.

\textbf{Graph Neural Networks.} 
Neural network model~\cite{scarselli2009graph} has been studied for years to preserve the information of a graph into vector space. Based on the graph spectral theory, graph convolutional networks have been proposed to capture the proximity information of the graph-structured data~\cite{DefferrardBV16, KipfW17}. To deal with the large scale real-world data, the spatial-based GNNs are proposed with a message passing strategy, called graph convolution. Specifically, the graph convolution learns the node representations by aggregating features from its neighbors. With $k$ times of graph convolution, the $k$-hop neighborhood information of nodes are encoded into feature vectors. Recently, various strategies have been proposed to advance the message passing in GNNs, including the advanced aggregation functions~\cite{HamiltonYL17}, node attention mechanisms~\cite{VelickovicCCRLB18}, graph structure poolings~\cite{gao2019graph}, neighborhood sampling methods~\cite{ChenMX18}, graph recurrent networks~\cite{huang2019graph}, and multi-channel GNNs~\cite{zhou2019multi}. In this work, we explore an orthogonal direction that learns a meta-policy to customize message passing strategy for each node in the graph.


\textbf{Graph Neural Architecture Search.} Given a specific task, neural architecture search (NAS) aims to discover the optimal model without laborious neural architecture tuning from the predefined architecture search space~\cite{zoph2016neural,pham2018efficient,jin2018auto, miikkulainen2019evolving,li2020pyodds}. 
Following the success of NAS, the concept of automation has been extended to GNNs architectures\cite{gao2019graphnas,zhou2019auto}.
The search space consists of all the possible variants of GNN architecture. A controller is then designed to explore the search space to find the optimal model and maximize the performance. The concept of meta-policy in \emph{Policy-GNN} is similar to the controller in NAS. Instead of focusing on searching the space of neural architecture, our meta-policy learns to sample the best iterations of aggregation for each node, and achieves much better performance in real-world benchmarks.



\textbf{Meta-policy Learning.} Recent advances in deep reinforcement learning (RL) have shown its wide applicability, such as games~\cite{MnihKSGAWR13,zha2019rlcard} and neural architecture search~\cite{zoph2016neural}. In general, the policy of deep neural networks is used to extract features from observations and make the corresponding decisions. The idea of meta-policy learning~\cite{zha2019experience} is to train a meta-policy by exploiting the feedback from the model to optimize the performance. The meta-policy learning has been widely studied to generate and replay experience data~\cite{zha2019experience}, to learn a dual policy for improving the sample efficiency of the agent~\cite{lai2020dual} and to maximize the performance by learning reward and discount factor~\cite{xu2018meta}. However, the successes of these studies are limited in simulated environments, which are far from real-world applications. Our work demonstrates the effectiveness of meta-policy learning in the context of GNNs in real-world scenario.

\section{Conclusions and Future work}
We present \emph{Policy-GNN}, a meta-policy framework that adaptively learns an aggregation policy to sample diverse iterations of aggregations for different nodes. Policy-GNN formulates the graph representation learning problem as a Markov decision process and employs deep reinforcement learning with a tailored reward function to train the meta-policy. To accelerate the learning process, we design a buffer mechanism to enable batch training and introduce parameter sharing to decrease the training cost. The conducted experiments on three real-world benchmark datasets suggest that \emph{Policy-GNN} significantly and consistently outperforms the state-of-the-art alternatives across all the datasets. Our work shows the promise in learning an aggregation strategy for GNNs, potentially opening a new avenue for future research in boosting the performance of GNNs with adaptive sampling and aggregation. 

A known limitation of \emph{Policy-GNN} is training efficiency. To learn an effective meta-policy, we need hundreds of training iterations, which could be problematic for large graphs. In the future, we are particularly interested in the possibility of developing more scalable and efficient meta-policy algorithms for large graphs. Moreover, we will explore different deep reinforcement learning algorithms and more advanced aggregation functions to further improve the performance. Another interesting direction is combining our meta-policy with the neural architecture search to jointly optimize the neural architecture and the aggregation strategy.

\vspace{-7.5pt}
\begin{acks}
The work is, in part, supported by NSF (\#IIS-1718840, \#IIS-1750074). The views and conclusions in this paper are those of the authors and should not be interpreted as representing any funding agencies.
\end{acks}
\vspace{-7.5pt}

\bibliography{paper}
\bibliographystyle{ACM-Reference-Format}
\clearpage

\appendix
\section*{Appendix: Reproducibility}
\section{Implementation Detail}
\label{appendix}
To develop the Policy-GNN, we implement two main components, GNN module and meta-policy module using Python as the programming language. The code will be released for reproducibility.

\subsection{Implementing GNN Module}
We implement the GCN for GNN module with one of the most popular open-source package for graph neural network, PyTorch-geometric~\cite{Fey/Lenssen/2019} (\url{https://github.com/rusty1s/pytorch_geometric}). We develop the GCN (\url{https://github.com/rusty1s/pytorch_geometric/blob/master/examples/gcn.py}) into a openAI-gym (\url{https://github.com/openai/gym/}) alike learnable environment (class), the class mainly includes 6 member variables: the dataset, learning rate, weight decay rate for Adam optimizer, max action (layer) number, batch size, and the meta-policy function (DQN). Whenever we initialize the environment, we initialize the maximum number of GNN layers and store them into a list with order. There are three main member functions in the GCN environment: \texttt{step}, \texttt{evaluate} and \texttt{test}. The \texttt{step} function is developed to train the GCN layers. The input of the function is an action for the current state and the output of the function is the next state and the reward. The \texttt{evaluate} function is developed to test the current performance on validation dataset. The \texttt{test} function is developed for testing the performance on test datasets.

In the \texttt{step} function, we first stack the GCN layers and train the layers as mentioned in Algorithm~\ref{alg:buffer}, then call the \texttt{evaluate} function to get the performance on validation dataset. Once we obtain the performance on validation dataset, we shape the reward signal mentioned in Section~\ref{sec:method} and store the performance into a historical list for future reward shaping. After that, we perform k-hop neighborhood sampling mentioned in Section~\ref{sec:method}, to obtain the attributes of the sampled neighbor as the next state and return the reward signal as the output.

In both \texttt{evaluate} and \texttt{test} function, we first obtain the nodes attributes of the validation/test dataset. In the beginning we utilize the meta-policy to decide the number of layers that each node requires. Then, we apply the buffer mechanism to separate the node attributes and actions. After that, we stack the trained GCN layers based on each action to evaluate on the validation/test dataset. 

\subsection{Implementing Meta-Policy Module}
We implement the deep Q-learning for meta-policy module based on an open-source package of reinforcement learning in game artificial intelligence~\cite{zha2019rlcard} (\url{https://github.com/datamllab/rlcard/blob/master/rlcard/agents/dqn_agent_pytorch.py}). The DQN meta-policy is developed as a class, which mainly includes 3 member variables: memory buffer, Q estimator target estimator. Memory buffer is a fixed size queue for storing past experiences. Q estimator is a multi-layer perceptron whose input is state and action, and the output is an estimated Q value for decision making. Target estimator is exactly the same as Q estimator. The Q estimator is used for decision making, whereas the target estimator is used for stabilizing the learning process. The idea of target estimator is to periodically copy the parameters from Q estimator to target estimator to stabilize the Q value estimations. In this way, the target estimator can be deemed as a slower version of Q estimator, and therefore can be applied to the second term of Bellman equation in Equation~\ref{eq:qlearning} for calculating the discounted Q value. More details can be found in the original papers~\cite{MnihKSRVBGRFOPB15}.

There are mainly four member functions in DQN meta-policy class: \texttt{Feed}, \texttt{Train}, \texttt{Predict} and \texttt{Learn}. The \texttt{Feed} function is used for feeding the past experiences into the memory buffer. The input of this function is a transition composed of current state, current action, next state and current reward. The \texttt{Train} function is developed for training the Q estimator. In the \texttt{train} function, the Q estimator is updated through Equation~\ref{eq:qlearning}, and the target estimator periodically copy the parameters from Q estimator. \texttt{Predict} function is developed for predicting the action based on the learned Q estimator, which is the meta-policy function $\widetilde{\pi}$ in the paper. The input of the \texttt{Predict} function is the state, and the output of the function is the action. In the \texttt{Predict} function, we apply epsilon greedy method, which chooses an action based on the Q estimator with $1 - \epsilon$ probability, and randomly chooses an action with $\epsilon$ probability. \texttt{Learn} function is the main function to learn the meta-policy where the input of the function is the environment (GCN) and the training timestep. In the  \texttt{Learn} function, we reset the input environment to get the initial state, then we generate several trajectories and call \texttt{Feed} function to store them to the memory buffer. In the end we call the \texttt{Train} function to train the Q estimator.

\subsection{Training the Policy-GNN}
To train the proposed framework, we first initialize a GCN environment with hyper-parameters of GCN, and a DQN meta-policy with hyper-parameters of DQN. Then, we iteratively train the meta-policy and GCN by calling the \texttt{Learn} function of meta-policy with input GCN environment. In the training process, we record the best performance on validation dataset, and test the performance on test data if we obtain a better performance on validation dataset.

\section{More Details}
\label{appendix:B}
In this section we provide more details about experiments and hyper-parameter setting for Policy-GNN.

\subsection{Details of Datasets and Baselines}
As described in Section~\ref{subsec:data}, we use the public available benchmark datasets (\url{https://github.com/kimiyoung/planetoid/raw/master/data}) which are widely used in GNN researches. Specifically, we follow the setting of all of the baselines, which are available in the open-source package PyTorch-geometric (\url{https://github.com/rusty1s/pytorch_geometric/blob/master/torch_geometric/datasets/planetoid.py}). For the baseline results, since we strictly follow the settings of all of our baselines, we directly report the classification performances of all of the baselines on the corresponding papers.

\subsection{Detailed Settings for Policy-GNN}
For the hyper-parameter settings of Policy-GNN, we elaborate the GNN module and the meta-policy module separately. For GNN module, we set the maximum number of layers as 5 and batch size as 128. For GCN layers, we use $relu$ as activation function and include dropout layers within every two GCN layers with dropout rate 0.5. To optimize the GNN, we use Adam optimizer with learning rate as 0.01 and weight decay rate as 0.0005. For meta-policy module, a (32, 64, 128, 64, 32) MLP for Q estimator and target estimator, and use a memory buffer with size 10000. We copy the target estimator every 1000 training steps, set the discount factor for Bellman equation as 0.95. The training batch size is 128 and the $\epsilon$ is generated by a linear scheduler which linearly decades the $\epsilon$ every 10 steps with start $1.0$ and end at $0.1$. The learning rate for training meta-policy is set to 0.0005. Furthermore, we also exploit a state normalizer to normalize the state representation with the running mean value and standard deviation. The input state $s$ will be normalized as $\frac{s - mean}{std}$, where mean and standard deviation is calculated by the most recent 1000 states. Note that normalizer is a common trick used in deep reinforcement learning for stabilizing the training.



\end{document}